\documentclass{article}

\usepackage{arxiv}

\usepackage[utf8]{inputenc} % allow utf-8 input
\usepackage[T1]{fontenc}    % use 8-bit T1 fonts
\usepackage{hyperref}       % hyperlinks
\usepackage{url}            % simple URL typesetting
\usepackage{booktabs}       % professional-quality tables
\usepackage{amsfonts}       % blackboard math symbols
\usepackage{nicefrac}       % compact symbols for 1/2, etc.
\usepackage{microtype}      % microtypography
\usepackage{lipsum}		% Can be removed after putting your text content
\usepackage{graphicx}
\usepackage{natbib}
\usepackage{doi}

\usepackage{comment}
\usepackage{arydshln}
\usepackage{soul}
\usepackage{enumitem}

\title{\textit{PSentScore}: Evaluating Sentiment Polarity in Dialogue Summarization}

\date{} 					% Or removing it

\author{ 
	{Yongxin ZHOU, Fabien RINGEVAL, and François PORTET} \\
Univ. Grenoble Alpes, CNRS, Inria, Grenoble INP, LIG, 38000 Grenoble, France \\
prenom.nom@univ-grenoble-alpes.fr \\
}

\hypersetup{
pdftitle={\textit{PSentScore}: Evaluating Sentiment Polarity in Dialogue Summarization},
pdfauthor={Yongxin ZHOU, Fabien RINGEVAL, and François PORTET},
pdfkeywords={Dialogue summarization, Affective computing, Evaluation},
}

\newcommand{\tokendict}{\texttt{token-dict.}}
\newcommand{\bertsst}{\texttt{BERT-SST3}}
\newcommand{\bertdssst}{\texttt{BERT-DS-SST3}}

\newcommand{\baseline}{\texttt{baseline-BART$_{Large}$}}
\newcommand{\baselinesubsampled}{\texttt{baseline\_sub-sampled}}
\newcommand{\baselinefiltered}{\texttt{baseline\_Filtered}}

\begin{document}
\maketitle

\begin{abstract}
Automatic dialogue summarization is a well-established task with the goal of distilling the most crucial information from human conversations into concise textual summaries. However, most existing research has predominantly focused on summarizing factual information, neglecting the affective content, which can hold valuable insights for analyzing, monitoring, or facilitating human interactions. In this paper, we introduce and assess a set of measures \textit{PSentScore}, aimed at quantifying the preservation of affective content in dialogue summaries.
Our findings indicate that state-of-the-art summarization models do not preserve well the affective content within their summaries. Moreover, we demonstrate that a careful selection of the training set for dialogue samples can lead to improved preservation of affective content in the generated summaries, albeit with a minor reduction in content-related metrics.
\end{abstract}

% keywords can be removed
\keywords{Dialogue summarization \and Evaluation \and Sentiment Polarity}

\section{Introduction}
\label{intro}
Automatic dialogue summarization has been widely studied and applied to various domains, including meeting \citep{AMICorpus, zhong-etal-2021-qmsum}, chat \citep{gliwa-etal-2019-samsum}, email thread \citep{zhang-etal-2021-emailsum}, media interview \citep{zhu-etal-2021-mediasum}, customer service \citep{favre-etal-2015-call, lin-etal-2021-csds} and medical dialogue \citep{song-etal-2020-summarizing}. 
However, most research has focused on summarizing factual information, leaving aside affective content.

While it is essential to summarize the most pertinent factual information, the subjective content can also provide valuable insights.
The integration of subjective content, such as affective aspects, into summaries could bring various benefits. These benefits encompass enhancing customer service, facilitating collaborative interactions, and offering improved support to healthcare patients. For instance, in the customer service sector, call center telephone conversations play a vital role in monitoring and enhancing service quality. Many calls contain emotional information that are deemed important to report \citep{Roman2008}.

Even though summarizing the affective part of dialogues could be highly valuable in many applications, it has been understudied, with only one study focusing on this topic \citep{Roman2008}. 
Affective content has been the target of a few summarization tasks such as opinion summarization \citep{wang-ling-2016-neural}. However, such tasks mainly focus on non-dialogue text reviews.
In dialogues, factual information and subjective content are often intertwined. Therefore, when summarizing dialogues, it remains crucial to capture and synthesize not only the objective facts but also the subjective content.

One of the main limitations to incorporating sentiment into summarization lies in the human guidelines used to write the human references. Often, summarization tasks are crafted with a primary focus on facts and objectives, providing little guidance to human summarizers on how to handle affective content, as noted by previous research \citep{Roman2008}. 
A recent counter-example is the DialogSum dataset \citep{chen-etal-2021-dialogsum}, where annotators were required to pay extra attention to several different aspects including \textit{Emotions}. This shift in dataset design illustrates a growing recognition within the research community of the importance of incorporating affective content into summaries.

In the context of dialogue summarization, a pertinent question is how to reliably measure the ability of generative models to capture and convey affective information from the input dialogue in generated summaries.
The current automatic evaluation methods for dialogue summarization mostly rely on $n$-gram comparisons and embedding distances between generated and reference summaries, while some studies proposed new metrics to evaluate faithfulness in dialogue summarization \citep{wang-etal-2022-analyzing}.
However, these metrics have been designed for factual correctness, and do not focus on evaluating the relevance of the affective content. 

In this paper we make contributions to the field of dialogue summarization, which are outlined below:\footnote{\url{https://github.com/yongxin2020/PSentScore}}

\begin{enumerate}%[noitemsep]
    \item We emphasize the importance of affective content for dialogue summarization, especially in the context of customer service and health care.
    \item We propose \textit{PSent}, a measure that calculates the proportion of affect charged words (positive/negative) in a given text.
    \item We built several systems running sentiment analysis at the word-level in dialogues and evaluated how much affective content is preserved in summaries using \textit{PSentScore} which is based on \textit{PSent}, a reference-less measure.
    \item We exploited the DialogSum Challenge framework to provide a reliable set of data and state-of-the-art methods for the automatic generation of summaries, and analyzed the affective content using \textit{PSentScore}.
\end{enumerate}

The results show that by filtering dialogues according to sentiment, we can significantly improve the preservation of both positive and negative sentiments in summarization, while preserving the performance of factual information. 

\section{Related Work}

In the context of describing the human state of mind, several terms are usually employed, such as affect, feeling, emotion, sentiment, and opinion, which are sometimes used interchangeably despite existing differences between them \citep{6797872}. 
In our study, even though the conversations are discourses, as we are performing a textual analysis, we consider affective content as sentiments expressed by the interlocutors and use the term \textit{sentiment} throughout the paper.

\subsection{Sentiment in Dialogue Summarization}

Reporting the affective states of interlocutors in dialogue summarization is important in several cases. For instance, it can improve the customer experience by finding out whether the customer feedback is positive or negative \citep{zhou-etal-2022-effectiveness}.
In the context of health care, it is crucial to know how patients feel during human interactions such as clinical meetings, or human-machine interactions such as digital therapies \citep{10.1007/978-3-030-80285-1_55}.

As we mentioned earlier, factual information and subjective content are often intertwined in dialogues, and while it is important to summarize the most relevant factual information, the subjective content can also provide key information. A study by \citet{Roman2008} revealed that whenever a dialogue contains an extreme emotion, this behavior is reported in human written dialogue summary. The study also shows that the emotional reporting varies considerably depending on the summarizer’s viewpoint, and that size constraints have no impact on the emotional content reported in the summaries. In addition to this empirical evidence, there are some theoretical arguments in favor of the presence of emotions/sentiment in dialogue summarization. 
For instance, \citet{tuggener-etal-2021-summarizing} mapped dialogue types (categorization of dialogue types according to \citealp{Walton1995-WALCID}) to summary items, and \textit{Emotions} was explicitly mapped and emphasized as one of the summary items along with the following dialogue types: \textit{Deliberation}, \textit{Information seeking}, and \textit{Eristics}. 

Despite theoretical and empirical support for the inclusion of affective information in dialogue summaries, the inclusion of emotions/sentiment as a summary item is not a common practice when designing datasets. In a recent survey \citep{tuggener-etal-2021-summarizing}, of the datasets that were listed, only one dataset -- Call Centre Conversation Summarization (CCCS) \citep{favre-etal-2015-call} -- was found to have exploited \textit{Emotions/Sentiment} as a summary item, while there is no mention of \textit{Emotions/Sentiment} in the guideline. 
It seems thus that research on sentiment in dialogue summarization suffers a lack of resources. We argue that this lack of development is mainly due to two reasons: 1) the fact that current corpora did not consider affect in their guideline for writing reference summary; and 2) that there is, to the best of our knowledge, no automatic measure to assess the affective aspect of a summary with respect to its original source. 

\subsection{Dialogue Summarization Corpora and Sentiment}\label{subsec:ds_datasets}

\begin{table*}[ht]
\centering
\footnotesize
\begin{tabular}{p{2.2cm}p{2.3cm}ccp{6.4cm}}
\hline
\textbf{Name} & \textbf{Domain} & \textbf{Language} & \textbf{Guideline} & \textbf{Guideline criteria for writing the reference summaries} \\ 
 &  &  & \textbf{Available} &  \\ 
        \hline
        AMI \citep{AMICorpus} & Meeting & English & Yes & 
        Abstractive summaries should have the following structure: abstract, decisions, problems/issues, actions.
        Extractive summaries: identify extracts from the transcript which jointly convey the correct kind of information about the meeting to fit the required purpose.
        The instructions do not mention emotion/sentiment. \\
        \hline
        RATP-DECODA \citep{favre-etal-2015-call} & Telephone Customer Service & French & No & We contacted the authors and obtained their summary definition, there is no mention of emotion/sentiment. \\ 
        \hline
        SAMSum \citep{gliwa-etal-2019-samsum} & Chat & English & Yes & (1) Be rather short, (2) extract important pieces of information, (3) include names of interlocutors, (4) be written in the third person. The  instructions do not mention emotion/sentiment.\\
        \hline
        MEDIASUM \citep{zhu-etal-2021-mediasum} & Media Interview & English & No & The reference summaries were downloaded from text descriptions of the input documents (interviews) available on the web.
        \\
        \hline
        TWEETSUMM \citep{feigenblat-etal-2021-tweetsumm-dialog} & Customer Service & English & Yes & Extractive summary: highlight the most salient sentences in the dialog. Abstractive summaries: one sentence summarizing what the customer conveyed and a second sentence summarizing what the agent responded. The instructions do not mention emotion/sentiment. \\
        \hline
        QMSum \citep{zhong-etal-2021-qmsum} & Multi-domain Meeting & English & Yes & The annotation process consists of three stages: topic segmentation, query generation, and query-based summarization. The  instructions do not mention emotion/sentiment.  \\
        \hline
        CSDS \citep{lin-etal-2021-csds} & Customer Service & Chinese & No & There are three different summaries for each dialogue: an overall summary and two role-oriented summaries (user and agent). Emotion/sentiment is not mentioned.
        \\
        \hline
        DIALOGSUM \citep{chen-etal-2021-dialogsum} & Spoken & English & Yes & Convey the most salient information; Be brief; Preserve important named entities within the conversation; Be written from an observer perspective; Be written in formal language. Pay extra attention to the following aspects: Tense Consistency, Discourse Relation, \textbf{Emotion} and Intent Identification. \\ 
        \hline
\end{tabular}
\caption{Major datasets for dialogue summarization with their summaries criteria. DialogSum is the only one to include \textit{Emotion} in the guideline.}
\label{tab1:DS_datasets}
\end{table*}

Despite dialogue summarization being a well-established task, the formulation of summarization tasks has not reached a consensus in the linguistic and the Natural Language Processing (NLP) communities, which has prevented from reaching a mutually agreed-upon definition of what a dialogue summary should look like \citep{guo-etal-2022-questioning}. 
In order to evaluate to which extent corpora used for dialogue summarization considered affective information in their summary, we performed a short overview of the guideline of several major dialogue summarization datasets, which have been widely used in NLP research. This is summarized in Table~\ref{tab1:DS_datasets}. For each corpus, we can see that the summary criteria used are different. For some corpora, only the data has been made available, while annotation guidelines are rarely accessible.

As it can be seen, some corpora do not reveal the criteria used for 
reference summaries. However, most of the corpora disclose their objectives to write reference summaries as for the AMI meeting corpus \citep{AMICorpus}, SAMSum \citep{gliwa-etal-2019-samsum} (written online conversation), TWEETSUMM \citep{feigenblat-etal-2021-tweetsumm-dialog}, which is focused on customer service, QMSum \citep{zhong-etal-2021-qmsum}, a query-based multi-domain meeting summarization dataset,
and DialogSum \citep{chen-etal-2021-dialogsum}, which is a real-life scenario dialogue summarization dataset. We can notice that amongst all those corpora, only in the DialogSum dataset, annotators were explicitly instructed to describe important emotions related to events in the reference summary.

The fact that the annotation is not explicitly tasked with processing sentiment does not prevent the reference summaries from containing it, to a certain extent.
We checked this by manually analyzing the 212 annotated synopses of 100 dialogues taken from the RATP DECODA corpus test set \citep{favre-etal-2015-call}. We found that, although annotators were not explicitly instructed to indicate customer satisfaction in the synopsis, some annotators did mention customer feelings, but this only occurred in a few cases, namely 4\% of the synopses. 

\subsection{Evaluation of Affective Content in Dialogue Summarization} 

Most evaluations of summarization tasks still rely on $n$-gram base measure such as ROUGE \citep{lin-2004-rouge}. The F1 scores of ROUGE-1, ROUGE-2 and ROUGE-L are mainly reported, which measure word overlap, bi-gram overlap and longest common sequence between generated summaries and references. 

Other measures such as BERTScore embeddings \citep{Zhang*2020BERTScore:} have also emerged to provide a more subtle evaluation of similarities by taking context and semantic proximity into account. 

Recent works \citep{huang-etal-2020-achieved, 10.1162/tacl_a_00373} have already pointed out that these metrics do not  correlate equally with all kinds of human judgments. However, we are not aware of any automatic metric measuring sentiment adequacy with sources in dialogue summarization or other linguistic summarization task.

\section{Measuring the Affective Content of Summaries}

In this section, we present our method for measuring the affective content of summaries. 
Our hypotheses rely on the following: 
\begin{itemize}%[noitemsep]
    \item It is possible to measure automatically the presence of sentiment in texts with sufficient reliability, whether measured by dimensional polarities or categories, and at word or sentence level.
    %\item It is possible to measure emotional content of a text (whether it is through polarity, dimensions, or categories) in terms of proportion.
    \item The distribution of affective content in the summaries should be similar to that of the dialogue.
    % \item Ideally, the summaries should contain the same proportion, or respectively the same polarity, of sentiment as the input source texts. Therefore, a system summarizing a document could be evaluated by measuring how far the sentiment proportion, or respectively polarity, of the generated summary is from the sentiment proportion, or respectively the sentiment polarity, of the input text. In case of several documents, a unique measure can be derived. 
    \item Since affective and factual content might be interleaved, the measure might correlate with other context-based measures so it cannot replace them.   
\end{itemize}

\subsection{The PSent measure}

In order to investigate whether the sentiment polarities of the input dialogues are preserved in the corresponding summaries, we assess the variability of sentiments in both the input dialogue and their corresponding summaries.
Since our hypothesis is that the affective content of the input should be preserved in the output summary to a comparable extent, we propose to compute a ratio for the input and output affective content.

There are different resolutions in which sentiment can be calculated (document, paragraph, sentences, words). As a summary can be very short we assume that the word-level is the most adequate. 

After labeling positive and negative words in each sentence using word-level Sentiment Analysis (SA) models, the number of positive and negative words and the total number of words in the input dialogue and corresponding summaries can be counted. We then calculate the proportion of affect charged words in the whole dialogue and in the summary, respectively. The formula used for this calculation is as follows:

\begin{equation}
PSent = (PosN+NegN) / TotalN\label{PSent}
\end{equation}

In eq.~\ref{PSent}, \textit{PSent} represents the proportion of sentimentally charged words in the given texts. We use \textit{PSentDial}, and \textit{PSentSumm} to represent the \textit{PSent} in the input dialogue, and reference summaries (or generated summaries), respectively. Furthermore, we can also compute \textit{PSent$_{P}$} (resp. \textit{PSent$_{N}$}) to denote the proportion of affect charged positive words -- $PosN$ -- only (resp. negative words -- $NegN$ -- only) in the given texts.

\subsection{The PSentScore measure} %CorrPSent
\label{subsec:PSentScore}

Ideally, the summaries should mirror the sentiment proportion or polarity of the input texts. Therefore, the evaluation of a summarization system can be performed by quantifying the disparity between the sentiment proportion or polarity in the generated summary and that in the input text. For multiple documents, a unified measure can be derived.

To examine whether the sentiment polarities presented in the input dialogues and in the output summaries are equivalent, we first calculate \textit{PSentDial} and \textit{PSentSumm} for each dialogue-summary pair.

Then to explore the relationship strength between \textit{PSentDial} and \textit{PSentSumm} in various splits of dialogue summarization datasets, we compute and present \textit{PSentScore} using the following measures:
1) Spearman’s rank correlation coefficient ($r_s$) -- eq.~\ref{spearman}, which assesses the monotonic relationships between two variables \citep{Zar2005SpearmanRC}; 2) Concordance Correlation Coefficient (CCC) -- eq.~\ref{ccc}, which quantifies the similarity between two sets of data,  i.e. the trends between two variables; 3) Mean Absolute Error (MAE) -- eq.~\ref{mae}, calculates errors between two sets of values, it is also known as scale-dependent accuracy as it calculates error in observations taken on the same scale.

\begin{equation}
    r_s = 1 - \frac{6\sum d_i^2}{n(n^2 - 1)}\label{spearman}
\end{equation}

Where: 
$r_s$ represents the Spearman's rank correlation coefficient.
$d_i$ represents the differences between the ranks of corresponding data points in the two variables being compared.
$n$ is the number of data points.

\begin{equation}
CCC(x, y) = \frac{2\rho\sigma_x\sigma_y}{\sigma_x^2 + \sigma_y^2 + (\mu_x - \mu_y)^2}\label{ccc}
\end{equation}

Where:
$\rho$ represents the Pearson correlation coefficient between x and y values.
$\sigma_x$ and $\sigma_y$ represent the standard deviation of the x and y values.
% $\sigma_y$ represents the standard deviation of the y values.
$\mu_x$ and $\mu_y$ represent the mean of the x and y values.

\begin{equation}
MAE(y, \hat{y}) = \frac{1}{n} \sum_{i=1}^{n} |y_i - \hat{y}_i| \label{mae}
\end{equation}

Where:
$n$ is the number of data points. $y_i$ and $\hat{y}_i$ represent the values of the two variables for the $i$-th
 data point, respectively.

\subsection{Experimental design} 

We intend to show the effect of the measure empirically. Hence the first step of the method is to select a corpus with reference texts containing some affective items. Out of all the corpora mentioned in Section \ref{subsec:ds_datasets}, we selected DialogSum \citep{chen-etal-2021-dialogsum}, which is composed of social conversations that are often affect charged. As a reminder, annotators were explicitly instructed to describe important emotions related to events in the summary. In addition, DialogSum was used in a challenge, in which several teams participated and presented their results \citep{chen-etal-2022-dialogsum}. 

We then computed $PosN$ and $NegN$ at the word-by-word level, using state-of-the-art models based on BERT \citep{devlin-etal-2019-bert}. These models were evaluated on a separate corpus and used to evaluate to which extent DialogSum contains sentiment in its documents. 

Finally, using \textit{PSentScore} and standard measures, we evaluated to which extent the state-of-the-art models handle sentiment with DialogSum. We then proposed a method to select the training target by eliminating documents without affective content in the input dialogue and/or summary, and trained models on this filtered data to evaluate whether sentiment handling can be improved.

\section{Measuring Affective Content of Reference Summaries}

Initially, we adopted the \textit{opinion\_lexicon} \citep{10.1145/1014052.1014073} dictionary as the simplest approach for Sentiment Analysis (SA). This dictionary consists of two lists of positive and negative words; any word that is not positive or negative is thus labeled as neutral. 
However, this dictionary-based approach has some limitations, as the polarity of some words may vary depending on their context (e.g., the word ``kind"), and such differences cannot be distinguished by this dictionary-based approach. To overcome this limitation, we then explored contextual SA at the word level and considered training a SA model for this purpose.

\subsection{Training Word-level Sentiment Analysis Models}

\begin{table*}[htbp]
\centering
\footnotesize
\begin{tabular}{lllll}
\hline
 & overall\_accuracy & precision & recall & f1 \\
\hline
token-dict.   & 88.82            & 73.61          & 60.96        & 65.64    \\
\hline
BERT-SST3    & 97.87 ($\pm$0.06)           & 94.43 ($\pm$0.43)           & 94.07 ($\pm$0.38)        & 94.24 ($\pm$0.15)    \\
\hline
BERT-DS-SST3 & 97.96 ($\pm$0.04)            & 94.53 ($\pm$0.17)           & 94.39 ($\pm$0.20)        & 94.46 ($\pm$0.10)    \\
\hline
\end{tabular}
\caption{Performances in terms of accuracy, precision, recall, f1 (\%) on the test set of the SST-3 dataset, for different models: \tokendict{}, \bertsst{} and \bertdssst{}. Statistics are given in the following format: mean (standard deviation), based on three runs. Macro results for precision, recall and f1.}
\label{tab:results_token_classification_sst}
\end{table*}

\subsubsection{Corpus: Stanford Sentiment Treebank (SST)} \label{sst_corpus}

The SST dataset~\citep{socher-etal-2013-recursive} is the first corpus that provides fully labeled parse trees, enabling a complete analysis of the compositional effects of sentiment in language. 
This dataset has been extensively studied for binary single sentence sentiment classification (positive/negative) and fine-grained sentiment classification (five classes). 
Given the complete parse tree annotations, it presents an opportunity to adapt it for word level SA. To the best of our knowledge, this is the only dataset available for word level sentiment classification.

The SST dataset includes fine-grained sentiment labels for 215,154 phrases in the parse trees of 11,855 sentences. The partition statistics are presented in Table~\ref{tab:statistic_sst}.
In the following, we only focus on studying word level polarity. Hence, we pre-processed the original SST that was annotated with a 5-point Likert scale (``very negative", ``negative", ``neutral", ``positive" and ``very positive") into a 3-point Likert scale by simply merging ``very negative" into ``negative" and ``very positive" into ``positive"; this will be referred to as \textbf{SST3} in this paper.   

\begin{table}[htbp]
\centering
\footnotesize
\begin{tabular}{ll}
\hline
\textbf{Split} & \textbf{\# samples} \\
\hline
Training & 8544 \\
\hline
Validation & 1101 \\
\hline
Test & 2210 \\
\hline
\end{tabular}
\caption{Basic statistics for SST dataset.}
\label{tab:statistic_sst}
\end{table}

\subsubsection{Word-level SA Models}

We built three specific models to perform word-level SA. 
To train BERT-based word-level SA models, we adapted the training code provided by Hugging Face for token classification tasks.\footnote{\url{https://github.com/huggingface/transformers/tree/main/examples/pytorch/token-classification}}

\paragraph{token-dict.}
The dictionary-based classifier based on \textit{opinion\_lexicon} \citep{10.1145/1014052.1014073}, it relies on a list of positive and negative opinion or sentiment words in English (about 6,800 words).

\paragraph{BERT-SST3}
For the second model, we fine-tuned BERT \citep{devlin-etal-2019-bert} 
% (bert-base-uncased) 
using the preprocessed SST3 dataset with word-level annotation, where each word receives a label. We used a learning rate of $5\mathrm{e}{-05}$ for 3 epochs. The model with the lowest validation loss was selected for reporting results on the test set and for further use.

\paragraph{BERT-DS-SST3}
As the SST dataset is composed of movie reviews and is not specific to conversational setting, we used domain adaptation to familiarize our model with dialogue-specific characteristics. To do so, we automatically annotated the DialogSum training partition using the \tokendict{} (dictionary-based classifier, each word gets a label) because it is independent of the domain.
We then fine-tuned BERT on this annotated corpus with a learning rate of $2\mathrm{e}{-05}$ for 5 epochs, selecting the model with the lowest validation loss.

Next, we proceeded to further fine-tune the selected model on the training set of SST3, which was annotated by human annotators. We used a learning rate of $5\mathrm{e}{-05}$ for 3 epochs. The aim was to obtain a model adapted to the DialogSum dataset but trained with reliable annotation from SST dataset. We selected the model with the lowest validation loss for evaluation on the test set and for future use. Both the training and prediction were performed on a NVIDIA Quadro RTX 6000 GPU.

\subsubsection{Word-level Sentiment Analysis Models' Performance}

Table~\ref{tab:results_token_classification_sst} shows the results of the models on the SST3 test set.
We evaluate token classification performance using common metrics such as token-level precision, recall, and F1 score. Due to an imbalance in the number of neutral token labels in preprocessed SST3, we report ``macro" results for these metrics when evaluating hypotheses.

The lexicon-based dictionary (\tokendict{}) shows poor performance on the SST token classification task, while \bertdssst{} performs similarly to \bertsst{}. It seems that domain adaptation has not decreased the performance on SST3 and, on the contrary, has stabilized its performance with a lower standard deviation than that of \bertsst{}.

\subsection{Affective Representation of DialogSum}

In what follows, we focus on comparing and evaluating the affective representation in input dialogues and reference summaries from the DialogSum dataset. We employ \bertdssst{} to calculate the \textit{PSent} of each dialogue and its corresponding reference summaries. 

\begin{figure}[ht]
    \centering
    \includegraphics[width=.6\textwidth]{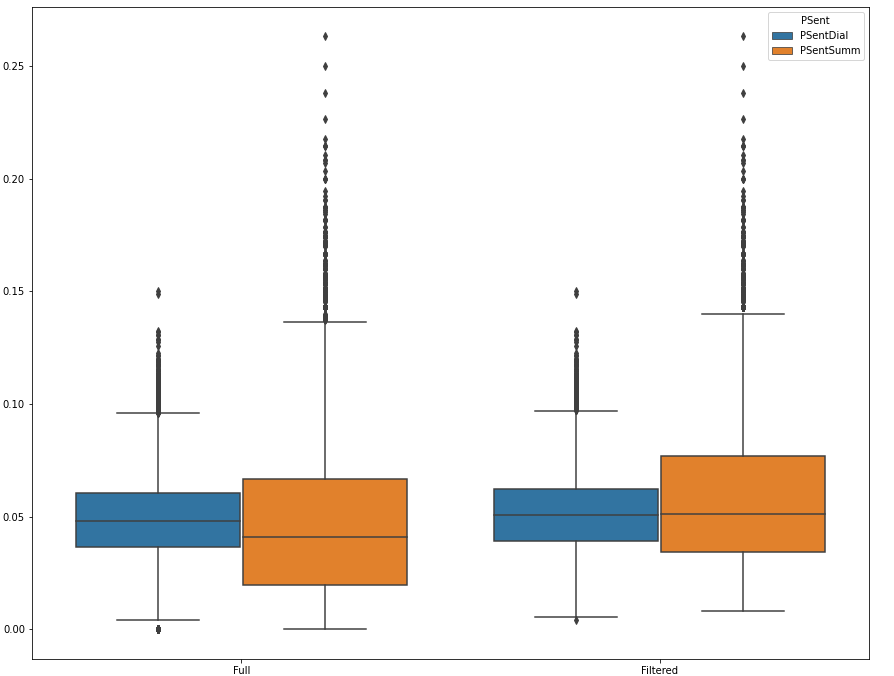}
    \caption{Box plots for \textit{PSentDial} (left) vs. \textit{PSentSumm} (right) distribution using \bertdssst{} on the full DialogSum training and validation sets. \textit{Filtered} means that samples with \textit{PSentDial} or \textit{PSentSumm} values equal to zero have been removed.
    \label{fig:PSent_BERT_DS_SST3}}
\end{figure}

In Figure~\ref{fig:PSent_BERT_DS_SST3}, we present box plots of the distributions of \textit{PSentDial} versus \textit{PSentSumm} for the DialogSum training and validation sets. The figure includes two versions of the distributions: \textit{Full}, which comprises all samples, and \textit{Filtered}, which removes samples with \textit{PSentDial} and/or \textit{PSentSumm} values equal to zero (cf. Table~\ref{tab: statistic_DialogSum} for the statistics). We use the \textit{Filtered} version to avoid the potential impact of zero values on reported results. 

Considering all samples, the median of \textit{PSentSumm} is lower than that of \textit{PSentDial}, indicating that there may be an under-representation of affective states in the reference summaries of the train and dev partitions, even though emotions are indicated in the corpus annotation guidelines. For the \textit{Filtered} distribution on the training and validation sets, the median of \textit{PSentSumm} is similar to that of \textit{PSentDial}. However, for both versions, the distributions outside of the quartiles of the box plots are more varied for \textit{PSentSumm} than for \textit{PSentDial}.

\section{Assessing Sentiment Handling of Summarization Models}

\subsection{Filtering Methodology}

In order to investigate the sentiment handling by the state-of-the-art models and to carefully select the training target by eliminating the pairs without affective content in the input dialogue or the summary, we filtered the DialogSum dataset using the word-level SA method mentioned earlier: \bertdssst{}.

\begin{table}[ht]
\centering
\footnotesize
\begin{tabular}{lllll}
\hline
Part.  & {Full}  & \textbf{Filtered} by & \multicolumn{2}{c}{\bf w/ zero value}     \\
           &                     & \textbf{BERT-DS-SST3} & Dial. & Sum. \\
\hline
train & 12460 & 9687 (77.5\%) & 43 & 2757  \\ 
dev & 500 & 391 (78.2\%)  & 2 & 108 \\
test & 500 & 499 (99.8\%)  & 1 & 0\\
\hline
\end{tabular}
\caption{\label{tab: statistic_DialogSum} Statistics for DialogSum dataset. For the Filtered corpus both input dialogue and reference summaries without affective content according to \bertdssst{} were removed. However, for the test partition only the input dialogues were filtered out. Percentage of data kept is shown in parentheses (\%).}
\end{table}

Detailed statistics for the DialogSum dataset are provided in Table~\ref{tab: statistic_DialogSum}, and \textit{Full set} represents its raw statistics: a total of 13,460 dialogues are divided into training (12,460), validation (500) and test (500) sets \citep{chen-etal-2021-dialogsum}. The \textit{Filtered} set contains 9687 training and 391 validation samples.

\subsection{Experimental Setup}

Following the state-of-the-art models \citep{chen-etal-2021-dialogsum},
% https://doi.org/10.48550/arxiv.2209.11910}
we fine-tuned the BART-Large model \citep{lewis-etal-2020-bart} on the full set of DialogSum and on the filtered dataset.\footnote{We adapted the training code from \citet{wang2022focused} to reproduce the results of the baseline model (BART-Large) on the DialogSum dataset.} 
We also trained the model on a corpus of the same size as the filtered dataset, but whose instances were randomly sampled from the full dataset.
The hyperparameters setting was 
learning rate of $5\mathrm{e}{-05}$ for 15 epochs. Experiments were performed on the NVIDIA Quadro RTX 6000 GPU and took about 2.5 hours for each run. 

\subsection{Evaluation Metrics}

In addition to ROUGE \citep{lin-2004-rouge}\footnote{We used the Hugging Face script of the ROUGE metric which uses the Google Research implementation \url{https://github.com/huggingface/datasets/blob/main/metrics/rouge/rouge.py}, which is also the one used in \citet{chen-etal-2021-dialogsum}.} and BERTScore \citep{Zhang*2020BERTScore:}\footnote{Following \citet{chen-etal-2022-dialogsum}, we use RoBERTa \citep{Liu2019RoBERTaAR} large as the backbone to compute BERTScore and the precision scores are reported.}, we propose a new set of measures to assess the relevance of a summary with respect to the affective charge (proportion - PSent), and its polarity (PSent$_{P}$ / PSent$_{N}$). We use \bertdssst{} as the backbone to calculate them. 
The evaluation methods from sentiment perspectives are as follows:

\textbf{PSentScore} is our proposed \textit{PSent} evaluation method from a proportional perspective. We compute the relationship strength between \textit{PSentDial} and \textit{PSentSumm} and provide Spearman/CCC/MAE scores. These scores indicate the monotonic relationships, trends, and errors between two sets of values, respectively (as mentioned earlier in \ref{subsec:PSentScore}). 

\textbf{PSentScore$_{P}$} and \textbf{PSentScore$_{N}$} are proposed from the polarity perspective. We examine whether the positive (resp. negative) affective aspects presented in the input dialogues are also present in the output summaries. 

For the above measures from affective perspectives, samples with zero \textit{PSentDial} (or \textit{PSent$_{P}$Dial} / \textit{PSent$_{N}$Dial}) values are removed, to account for the potential impact of zero values on the reported correlation. 

\subsection{Quantitative Results} 
\begin{table*}
\centering
\resizebox{\textwidth}{!}{
\begin{tabular}{l|cccc|ccc}
\hline
\textbf{Model} & \textbf{R1} & \textbf{R2} & \textbf{RL} & \textbf{BERTScore}  & \textbf{PSentScore} & \textbf{PSentScore$_P$}  & \textbf{PSentScore$_N$}\\
\# samples & 500* & 500* & 500* & 500* & 499$\dagger$ & 491$\dagger$ & 419$\dagger$ \\
\hline
Human \citep{chen-etal-2022-dialogsum} & 53.35  & 26.72 & 50.84  & 92.63  & - &  -   & - \\
\hline
BART$_{Large}$ \citep{chen-etal-2021-dialogsum} & 47.28  & 21.18  & 44.83   & - &  -   & - & - \\
GoodBai \citep{chen-etal-2022-dialogsum} & {\bf 47.61} & {\bf 21.66} & 45.48  & {\bf 92.72} & .357/.289/.027  & .341/.307/.024  & .397/.358/\textbf{.014} \\
UoT \citep{lundberg-etal-2022-dialogue} & 47.29 & 21.65   & {\bf 45.92}  & 92.26 &    .356/.297/.027     & .364/.325/.023 & .383/.338/.014\\
IITP-CUNI \citep{bhattacharjee-etal-2022-multi} & 47.26  & 21.18  & 45.17   & 92.70 &  .348/.289/.031  & .311/.280/.027 & .397/.295/.018 \\
TCS\_WITM \citep{chauhan-etal-2022-tcs} & 47.02  & 21.20   & 44.90  & 90.13  & .364/.294/.028 &  {\bf .375}/.331/.024 & .431/.333/.014 \\ 
\hline
\textbf{baseline-BART$_{Large}$}   & 47.36      & 21.23   & 44.88   &  91.42    &  .353/.292/.029 &   .318/.295/.025    & .395/.322/.016 \\
\textbf{baseline\_sub-sampled}$\diamond$  & 46.94   & 20.52  & 44.43     & 91.29 &  .351/.294/.028 & .351/.319/.024 & .410/.352/.016 \\
\textbf{baseline\_Filtered}$\diamond$  & 45.78 & 19.69  & 43.21  & 90.83 &   \textbf{.435}/\textbf{.348}/\textbf{.027} & .370/\textbf{.352}/\textbf{.023} & {\bf .449}/\textbf{.373}/.015 \\
\hline
\end{tabular}
}
\caption{\label{tab: dialogsum_results} Comparison of results from the DialogSum challenge teams and our BART-Large models fine-tuned on the full (\baseline{}) and filtered corpus (\baselinefiltered{}) of the DialogSum dataset. The \baselinesubsampled{} model has been trained on a corpus of the same size as the Filtered dataset but whose instances have been randomly sampled from the the full DialogSum dataset. $\diamond$ indicates training on partial corpora: training set 9687 (77.5\%), dev set 391 (78.2\%). \textit{PSentScore} values indicate evaluation results: Spearman (↑) / CCC (↑) / MAE (↓). * refers to the Full DialogSum test set, and $\dagger$ to the Filtered test set.}
\end{table*}

\begin{figure*}[h!]
    \centering
    \includegraphics[width=\textwidth]{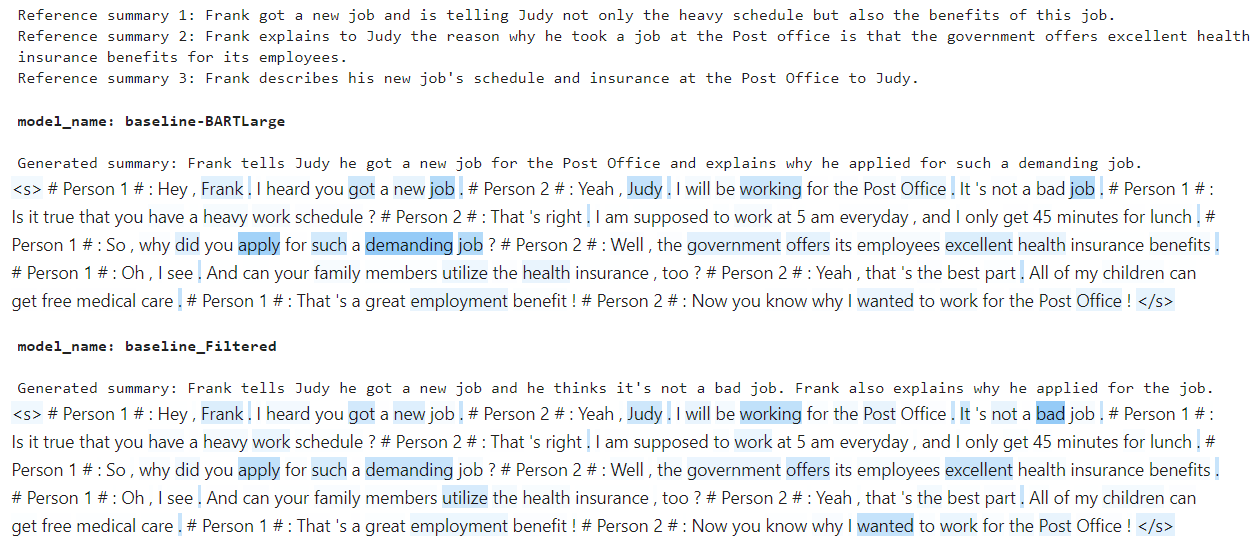}
    \caption{Example of test\_20, with three references, and predictions as well as visualization of the attention of two models: \baseline{} and \baselinefiltered{}.}
    \label{fig:test_20}
\end{figure*}

The results of the fine-tuned BART-Large model on different versions of the DialogSum dataset are presented in Table~\ref{tab: dialogsum_results}. We compare our results with previous results reported in DialogSum dataset paper \citep{chen-etal-2021-dialogsum}, and with results from different teams in the challenge \citep{chen-etal-2022-dialogsum, lundberg-etal-2022-dialogue, bhattacharjee-etal-2022-multi, chauhan-etal-2022-tcs}.
The predictions of the various systems were obtained from the corresponding authors, based on which we evaluated the \textit{PSentScore} results. \textit{Human} results are those computed by \citep{chen-etal-2022-dialogsum} obtained by averaging each human annotator scores against others. 

As reported in Table \ref{tab: dialogsum_results}, the GoodBai model \citep{chen-etal-2022-dialogsum} provides the highest ROUGE and BERT scores very close to the other teams and slightly better than the \texttt{BART$_{Large}$} model (2$^{nd}$ line) provided as reference to the challenge. Our \textbf{\baseline{}} model shows similar performances as the \texttt{BART$_{Large}$} model (2$^{nd}$ line). When trained on the Filtered dataset, the \textbf{\baselinefiltered{}} model exhibits a decrease of almost 1.5 points on all the ROUGE and BERT scores. However, when looking at the \textit{PSentScore} measures, the \textbf{\baselinefiltered{}} model provides the best correlation of affective content between dialogues and summaries (.435/.348/.027), far from the state-of-the-art models (.364/.297 in Spearman and CCC at most, while lowest MAE value -- .027 is similar).

Looking at \textit{PSentScore$_P$} and \textit{PSentScore$_N$},  the \textbf{\baselinefiltered{}} model also provides the best correlation of affective content in terms of polarity between dialogues and summaries (.370/.352/.023 and .449/.373/.015), while TCS\_WITM \citep{chauhan-etal-2022-tcs} reached .375 on \textit{PSentScore$_P$} in Spearman and GoodBai \citep{chen-etal-2022-dialogsum} reached .014 on \textit{PSentScore$_N$} in MAE, which are better than \textbf{\baselinefiltered{}}.

The \textbf{\baselinesubsampled{}} showed a decrease in ROUGE and BERTScore compared to the \textbf{\baseline{}} with a reduced number of training samples, while the \textit{PSentScore} measures were similar. 
The results show that by filtering dialogue samples according to affective content (\textbf{\baselinefiltered{}}), we can significantly increase the preservation of both positive and negative sentiment in summaries, while preserving factual information. It's worth noting that certain decreases in ROUGE and BERTScore may be due to the smaller amount of training data.

\subsection{Example Analysis}

To determine whether attention focuses more on affective words for our models trained with the filtered dataset, we visualized the distribution of attention weights for several examples. Examples are chosen with maximum \textit{PSentDial} using the word-level sentiment analysis models: \bertdssst{} and \tokendict{}.

We then visualized the distribution of their attention in the input dialogue for the following two models: \baseline{} and \baselinefiltered{}. In detail, we have calculated the encoder-decoder cross-attention, calculated the attention weights of the last layer (12$^{th}$) and the last head (16$^{th}$), the calculated attention weights are the dialogue versus the summary.

In what follows, we present one example with the predictions and visualization of attention from the two models mentioned previously. In Figure~\ref{fig:test_20}, the predicted summary of \baselinefiltered{} includes ``he thinks it's not a bad job", in addition to the factual information ``Frank tells Judy he got a new job" presented in both predictions, we can see that the word ``bad" in the dialogue is particularly emphasized, and the word ``excellent" receives more attention in the last model.

\section{Discussion and Conclusion}

In dialogue summarization, the most important content almost always focuses on factual information, leaving aside the affective content of the interaction. We argue that affective information is important content to report in dialogue summaries. In order to measure affective content in dialogues and in summaries, we trained SA models at the word level. We conducted a corpus-based analysis on the DialogSum corpus, in which dataset annotators were explicitly instructed to include affective content when writing reference summaries, we show that affective content is omitted to some extent in reference summaries in dialogue datasets. 

We then propose a new set of measures to evaluate the relevance of a summary based on the affective load (proportion) and its polarity (positive/negative).
Using this measure, we show that the summarization model often exhibits a mismatch between the affective content of the input dialogue and the summary. We also show that by carefully selecting the training target, we can decrease this mismatch. This method provides a more comprehensive measure of dialogue summarization performance.

In this study, we chose the DialogSum corpus for analysis because it explicitly considers emotions in its annotation guidelines. In the future, we will extend our method and conduct large-scale analyses on various dialogue summarization datasets and with more fine-grained affective categories. We also plan to extend the use of \textit{PSentScore} to other NLP tasks, such as summarizing reviews/opinions, generating emotional dialogues, etc.
We can also consider maximizing the \textit{PSentScore} for dialogue summarization as a functional goal, but this first requires an appropriate affect-oriented dataset.

\section*{Limitations}

Our measure is still gross and focuses on proportion and polarity perspectives, tested only on one data set. It does not distinguish, for example, whether we are reporting anger or sadness with the same distribution as in the dialogue. For this reason, we will look at other measures that might account for this.
Furthermore, the method currently only works for English.

We should also emphasize that the \textit{PSent} measure depends on a word-level sentiment analysis model which might not be available or biased if trained on a dataset different from the one \textit{PSent} is applied to. 
While our experiments focused on increasing the similarity of proportion of sentiment in the input and output texts, we did not perform a human evaluation of the outputs that might have provided more fine grained analyses. The standard automatic measures suggest that the summaries generated by the different models are somewhat similar but we recognize that the model learned on the Filtered corpus may generate degraded outputs due to less training data. Furthermore, the focus of the measure on the polarity is a crude evaluation of affective content that cannot account for subtle difference between the input text and the generated summaries as most of the other automatic measures.

While the \bertdssst{} model demonstrated promising performance on the SST corpus, it has not been evaluated on dialogue corpora. Annotating sentiment at the word level poses challenges, as annotators often lack consensus in their annotations. Our subsequent investigation will focus on sentiment analysis at the expression level. However, since \textit{PSent} measures the conservation and proportion of sentimental words, it remains a suitable metric for our purposes.

\section*{Ethics Statement}
The DialogSum corpus we used in this study is composed of resources freely available online without copyright constraint for academic use. According to the authors, the annotators had degrees in English Linguistics or Applied Linguistics. They received a salary of around
9.5 dollars per hour and took this annotation as a part-time job. We chose this corpus because it is the only dialogue summarization corpus we found that mentions emotions in its annotation guidelines, but we also acknowledge that the corpus consisting of social conversations may differ from other kinds of  conversations such as in the medical domain or customer service that may have specific characteristics.

\section{Acknowledgements}

This research was supported by the Banque Publique d'Investissement (BPI) under grant agreement THERADIA and was partially supported by MIAI@Grenoble-Alpes (ANR-19-P3IA-0003). We would also like to thank the anonymous reviewers for their insightful comments.

\bibliographystyle{unsrtnat}
\bibliography{references, anthology}  

\begin{thebibliography}{34}
\providecommand{\natexlab}[1]{#1}
\providecommand{\url}[1]{\texttt{#1}}
\expandafter\ifx\csname urlstyle\endcsname\relax
  \providecommand{\doi}[1]{doi: #1}\else
  \providecommand{\doi}{doi: \begingroup \urlstyle{rm}\Url}\fi

\bibitem[Carletta et~al.(2005)Carletta, Ashby, Bourban, Flynn, Guillemot, Hain,
  Kadlec, Karaiskos, Kraaij, Kronenthal, Lathoud, Lincoln, Lisowska, McCowan,
  Post, Reidsma, and Wellner]{AMICorpus}
Jean Carletta, Simone Ashby, Sebastien Bourban, Mike Flynn, Mael Guillemot,
  Thomas Hain, Jaroslav Kadlec, Vasilis Karaiskos, Wessel Kraaij, Melissa
  Kronenthal, Guillaume Lathoud, Mike Lincoln, Agnes Lisowska, Iain McCowan,
  Wilfried Post, Dennis Reidsma, and Pierre Wellner.
\newblock The ami meeting corpus: A pre-announcement.
\newblock In \emph{Proceedings of the Second International Conference on
  Machine Learning for Multimodal Interaction}, MLMI'05, page 28–39, Berlin,
  Heidelberg, 2005. Springer-Verlag.
\newblock ISBN 3540325492.
\newblock \doi{10.1007/11677482_3}.
\newblock URL \url{https://doi.org/10.1007/11677482\_3}.

\bibitem[Zhong et~al.(2021)Zhong, Yin, Yu, Zaidi, Mutuma, Jha, Awadallah,
  Celikyilmaz, Liu, Qiu, and Radev]{zhong-etal-2021-qmsum}
Ming Zhong, Da~Yin, Tao Yu, Ahmad Zaidi, Mutethia Mutuma, Rahul Jha,
  Ahmed~Hassan Awadallah, Asli Celikyilmaz, Yang Liu, Xipeng Qiu, and Dragomir
  Radev.
\newblock {QMS}um: A new benchmark for query-based multi-domain meeting
  summarization.
\newblock In \emph{Proceedings of the 2021 Conference of the North American
  Chapter of the Association for Computational Linguistics: Human Language
  Technologies}, pages 5905--5921, Online, June 2021. Association for
  Computational Linguistics.
\newblock \doi{10.18653/v1/2021.naacl-main.472}.
\newblock URL \url{https://aclanthology.org/2021.naacl-main.472}.

\bibitem[Gliwa et~al.(2019)Gliwa, Mochol, Biesek, and
  Wawer]{gliwa-etal-2019-samsum}
Bogdan Gliwa, Iwona Mochol, Maciej Biesek, and Aleksander Wawer.
\newblock {SAMS}um corpus: A human-annotated dialogue dataset for abstractive
  summarization.
\newblock In \emph{Proceedings of the 2nd Workshop on New Frontiers in
  Summarization}, pages 70--79, Hong Kong, China, November 2019. Association
  for Computational Linguistics.
\newblock \doi{10.18653/v1/D19-5409}.
\newblock URL \url{https://aclanthology.org/D19-5409}.

\bibitem[Zhang et~al.(2021)Zhang, Celikyilmaz, Gao, and
  Bansal]{zhang-etal-2021-emailsum}
Shiyue Zhang, Asli Celikyilmaz, Jianfeng Gao, and Mohit Bansal.
\newblock {E}mail{S}um: Abstractive email thread summarization.
\newblock In \emph{Proceedings of the 59th Annual Meeting of the Association
  for Computational Linguistics and the 11th International Joint Conference on
  Natural Language Processing (Volume 1: Long Papers)}, pages 6895--6909,
  Online, August 2021. Association for Computational Linguistics.
\newblock \doi{10.18653/v1/2021.acl-long.537}.
\newblock URL \url{https://aclanthology.org/2021.acl-long.537}.

\bibitem[Zhu et~al.(2021)Zhu, Liu, Mei, and Zeng]{zhu-etal-2021-mediasum}
Chenguang Zhu, Yang Liu, Jie Mei, and Michael Zeng.
\newblock {M}edia{S}um: A large-scale media interview dataset for dialogue
  summarization.
\newblock In \emph{Proceedings of the 2021 Conference of the North American
  Chapter of the Association for Computational Linguistics: Human Language
  Technologies}, pages 5927--5934, Online, June 2021. Association for
  Computational Linguistics.
\newblock \doi{10.18653/v1/2021.naacl-main.474}.
\newblock URL \url{https://aclanthology.org/2021.naacl-main.474}.

\bibitem[Favre et~al.(2015)Favre, Stepanov, Trione, B{\'e}chet, and
  Riccardi]{favre-etal-2015-call}
Benoit Favre, Evgeny Stepanov, J{\'e}r{\'e}my Trione, Fr{\'e}d{\'e}ric
  B{\'e}chet, and Giuseppe Riccardi.
\newblock Call centre conversation summarization: A pilot task at multiling
  2015.
\newblock In \emph{Proceedings of the 16th Annual Meeting of the Special
  Interest Group on Discourse and Dialogue}, pages 232--236, Prague, Czech
  Republic, September 2015. Association for Computational Linguistics.
\newblock \doi{10.18653/v1/W15-4633}.
\newblock URL \url{https://aclanthology.org/W15-4633}.

\bibitem[Lin et~al.(2021)Lin, Ma, Zhu, Xiang, Zhou, Zhang, and
  Zong]{lin-etal-2021-csds}
Haitao Lin, Liqun Ma, Junnan Zhu, Lu~Xiang, Yu~Zhou, Jiajun Zhang, and
  Chengqing Zong.
\newblock {CSDS}: A fine-grained {C}hinese dataset for customer service
  dialogue summarization.
\newblock In \emph{Proceedings of the 2021 Conference on Empirical Methods in
  Natural Language Processing}, pages 4436--4451, Online and Punta Cana,
  Dominican Republic, November 2021. Association for Computational Linguistics.
\newblock \doi{10.18653/v1/2021.emnlp-main.365}.
\newblock URL \url{https://aclanthology.org/2021.emnlp-main.365}.

\bibitem[Song et~al.(2020)Song, Tian, Wang, and
  Xia]{song-etal-2020-summarizing}
Yan Song, Yuanhe Tian, Nan Wang, and Fei Xia.
\newblock Summarizing medical conversations via identifying important
  utterances.
\newblock In \emph{Proceedings of the 28th International Conference on
  Computational Linguistics}, pages 717--729, Barcelona, Spain (Online),
  December 2020. International Committee on Computational Linguistics.
\newblock \doi{10.18653/v1/2020.coling-main.63}.
\newblock URL \url{https://aclanthology.org/2020.coling-main.63}.

\bibitem[Roman et~al.(2008)Roman, Piwek, and Carvalho]{Roman2008}
Norton Roman, Paul Piwek, and Ariadne Carvalho.
\newblock Emotion and behaviour in automatic dialogue summarisation.
\newblock 10 2008.
\newblock \doi{10.1145/1809980.1810056}.

\bibitem[Wang and Ling(2016)]{wang-ling-2016-neural}
Lu~Wang and Wang Ling.
\newblock Neural network-based abstract generation for opinions and arguments.
\newblock In \emph{Proceedings of the 2016 Conference of the North {A}merican
  Chapter of the Association for Computational Linguistics: Human Language
  Technologies}, pages 47--57, San Diego, California, June 2016. Association
  for Computational Linguistics.
\newblock \doi{10.18653/v1/N16-1007}.
\newblock URL \url{https://aclanthology.org/N16-1007}.

\bibitem[Chen et~al.(2021)Chen, Liu, Chen, and Zhang]{chen-etal-2021-dialogsum}
Yulong Chen, Yang Liu, Liang Chen, and Yue Zhang.
\newblock {D}ialog{S}um: {A} real-life scenario dialogue summarization dataset.
\newblock In \emph{Findings of the Association for Computational Linguistics:
  ACL-IJCNLP 2021}, pages 5062--5074, Online, August 2021. Association for
  Computational Linguistics.
\newblock \doi{10.18653/v1/2021.findings-acl.449}.
\newblock URL \url{https://aclanthology.org/2021.findings-acl.449}.

\bibitem[Wang et~al.(2022{\natexlab{a}})Wang, Zhang, Zhang, Chen, and
  Li]{wang-etal-2022-analyzing}
Bin Wang, Chen Zhang, Yan Zhang, Yiming Chen, and Haizhou Li.
\newblock Analyzing and evaluating faithfulness in dialogue summarization.
\newblock In \emph{Proceedings of the 2022 Conference on Empirical Methods in
  Natural Language Processing}, pages 4897--4908, Abu Dhabi, United Arab
  Emirates, December 2022{\natexlab{a}}. Association for Computational
  Linguistics.
\newblock URL \url{https://aclanthology.org/2022.emnlp-main.325}.

\bibitem[Munezero et~al.(2014)Munezero, Montero, Sutinen, and Pajunen]{6797872}
Myriam Munezero, Calkin~Suero Montero, Erkki Sutinen, and John Pajunen.
\newblock Are they different? affect, feeling, emotion, sentiment, and opinion
  detection in text.
\newblock \emph{IEEE Transactions on Affective Computing}, 5\penalty0
  (2):\penalty0 101--111, 2014.
\newblock \doi{10.1109/TAFFC.2014.2317187}.

\bibitem[Zhou et~al.(2022)Zhou, Portet, and
  Ringeval]{zhou-etal-2022-effectiveness}
Yongxin Zhou, Fran{\c{c}}ois Portet, and Fabien Ringeval.
\newblock Effectiveness of {F}rench language models on abstractive dialogue
  summarization task.
\newblock In \emph{Proceedings of the Thirteenth Language Resources and
  Evaluation Conference}, pages 3571--3581, Marseille, France, June 2022.
  European Language Resources Association.
\newblock URL \url{https://aclanthology.org/2022.lrec-1.382}.

\bibitem[Tarpin-Bernard et~al.(2021)Tarpin-Bernard, Fruitet, Vigne, Constant,
  Chainay, Koenig, Ringeval, Bouchot, Bailly, Portet, Alisamir, Zhou, Serre,
  Delerue, Fournier, Berenger, Zsoldos, Perrotin, Elisei, Lenglet, Puaux,
  Pacheco, Fouillen, and Ghenassia]{10.1007/978-3-030-80285-1_55}
Franck Tarpin-Bernard, Joan Fruitet, Jean-Philippe Vigne, Patrick Constant,
  Hanna Chainay, Olivier Koenig, Fabien Ringeval, B{\'e}atrice Bouchot,
  G{\'e}rard Bailly, Fran{\c{c}}ois Portet, Sina Alisamir, Yongxin Zhou, Jean
  Serre, Vincent Delerue, Hippolyte Fournier, K{\'e}vin Berenger, Isabella
  Zsoldos, Olivier Perrotin, Fr{\'e}d{\'e}ric Elisei, Martin Lenglet, Charles
  Puaux, L{\'e}o Pacheco, M{\'e}lodie Fouillen, and Didier Ghenassia.
\newblock Theradia: Digital therapies augmented by artificial intelligence.
\newblock In Hasan Ayaz, Umer Asgher, and Lucas Paletta, editors,
  \emph{Advances in Neuroergonomics and Cognitive Engineering}, pages 478--485,
  Cham, 2021. Springer International Publishing.
\newblock ISBN 978-3-030-80285-1.

\bibitem[Tuggener et~al.(2021)Tuggener, Mieskes, Deriu, and
  Cieliebak]{tuggener-etal-2021-summarizing}
Don Tuggener, Margot Mieskes, Jan Deriu, and Mark Cieliebak.
\newblock Are we summarizing the right way? a survey of dialogue summarization
  data sets.
\newblock In \emph{Proceedings of the Third Workshop on New Frontiers in
  Summarization}, pages 107--118, Online and in Dominican Republic, November
  2021. Association for Computational Linguistics.
\newblock \doi{10.18653/v1/2021.newsum-1.12}.
\newblock URL \url{https://aclanthology.org/2021.newsum-1.12}.

\bibitem[Walton and Krabbe(1995)]{Walton1995-WALCID}
Douglas~Neil Walton and Erik C.~W. Krabbe.
\newblock \emph{Commitment in Dialogue: Basic Concepts of Interpersonal
  Reasoning}.
\newblock Albany, NY, USA: State University of New York Press, 1995.

\bibitem[Feigenblat et~al.(2021)Feigenblat, Gunasekara, Sznajder, Joshi,
  Konopnicki, and Aharonov]{feigenblat-etal-2021-tweetsumm-dialog}
Guy Feigenblat, Chulaka Gunasekara, Benjamin Sznajder, Sachindra Joshi, David
  Konopnicki, and Ranit Aharonov.
\newblock {TWEETSUMM} - a dialog summarization dataset for customer service.
\newblock In \emph{Findings of the Association for Computational Linguistics:
  EMNLP 2021}, pages 245--260, Punta Cana, Dominican Republic, November 2021.
  Association for Computational Linguistics.
\newblock \doi{10.18653/v1/2021.findings-emnlp.24}.
\newblock URL \url{https://aclanthology.org/2021.findings-emnlp.24}.

\bibitem[Guo et~al.(2022)Guo, Clavel, Kamal~Eddine, and
  Vazirgiannis]{guo-etal-2022-questioning}
Yanzhu Guo, Chlo{\'e} Clavel, Moussa Kamal~Eddine, and Michalis Vazirgiannis.
\newblock Questioning the validity of summarization datasets and improving
  their factual consistency.
\newblock In \emph{Proceedings of the 2022 Conference on Empirical Methods in
  Natural Language Processing}, pages 5716--5727, Abu Dhabi, United Arab
  Emirates, December 2022. Association for Computational Linguistics.
\newblock URL \url{https://aclanthology.org/2022.emnlp-main.386}.

\bibitem[Lin(2004)]{lin-2004-rouge}
Chin-Yew Lin.
\newblock {ROUGE}: A package for automatic evaluation of summaries.
\newblock In \emph{Text Summarization Branches Out}, pages 74--81, Barcelona,
  Spain, July 2004. Association for Computational Linguistics.
\newblock URL \url{https://aclanthology.org/W04-1013}.

\bibitem[Zhang* et~al.(2020)Zhang*, Kishore*, Wu*, Weinberger, and
  Artzi]{Zhang*2020BERTScore:}
Tianyi Zhang*, Varsha Kishore*, Felix Wu*, Kilian~Q. Weinberger, and Yoav
  Artzi.
\newblock Bertscore: Evaluating text generation with bert.
\newblock In \emph{International Conference on Learning Representations}, 2020.
\newblock URL \url{https://openreview.net/forum?id=SkeHuCVFDr}.

\bibitem[Huang et~al.(2020)Huang, Cui, Yang, Bao, Wang, Xie, and
  Zhang]{huang-etal-2020-achieved}
Dandan Huang, Leyang Cui, Sen Yang, Guangsheng Bao, Kun Wang, Jun Xie, and Yue
  Zhang.
\newblock What have we achieved on text summarization?
\newblock In \emph{Proceedings of the 2020 Conference on Empirical Methods in
  Natural Language Processing (EMNLP)}, pages 446--469, Online, November 2020.
  Association for Computational Linguistics.
\newblock \doi{10.18653/v1/2020.emnlp-main.33}.
\newblock URL \url{https://aclanthology.org/2020.emnlp-main.33}.

\bibitem[Fabbri et~al.(2021)Fabbri, Kryściński, McCann, Xiong, Socher, and
  Radev]{10.1162/tacl_a_00373}
Alexander~R. Fabbri, Wojciech Kryściński, Bryan McCann, Caiming Xiong,
  Richard Socher, and Dragomir Radev.
\newblock {SummEval: Re-evaluating Summarization Evaluation}.
\newblock \emph{Transactions of the Association for Computational Linguistics},
  9:\penalty0 391--409, 04 2021.
\newblock ISSN 2307-387X.
\newblock \doi{10.1162/tacl_a_00373}.
\newblock URL \url{https://doi.org/10.1162/tacl\_a\_00373}.

\bibitem[Zar(2005)]{Zar2005SpearmanRC}
Jerrold~H. Zar.
\newblock Spearman rank correlation.
\newblock 2005.

\bibitem[Chen et~al.(2022)Chen, Deng, Liu, and Zhang]{chen-etal-2022-dialogsum}
Yulong Chen, Naihao Deng, Yang Liu, and Yue Zhang.
\newblock {D}ialog{S}um challenge: Results of the dialogue summarization shared
  task.
\newblock In \emph{Proceedings of the 15th International Conference on Natural
  Language Generation: Generation Challenges}, pages 94--103, Waterville,
  Maine, USA and virtual meeting, July 2022. Association for Computational
  Linguistics.
\newblock URL \url{https://aclanthology.org/2022.inlg-genchal.14}.

\bibitem[Devlin et~al.(2019)Devlin, Chang, Lee, and
  Toutanova]{devlin-etal-2019-bert}
Jacob Devlin, Ming-Wei Chang, Kenton Lee, and Kristina Toutanova.
\newblock {BERT}: Pre-training of deep bidirectional transformers for language
  understanding.
\newblock In \emph{Proceedings of the 2019 Conference of the North {A}merican
  Chapter of the Association for Computational Linguistics: Human Language
  Technologies, Volume 1 (Long and Short Papers)}, pages 4171--4186,
  Minneapolis, Minnesota, June 2019. Association for Computational Linguistics.
\newblock \doi{10.18653/v1/N19-1423}.
\newblock URL \url{https://aclanthology.org/N19-1423}.

\bibitem[Hu and Liu(2004)]{10.1145/1014052.1014073}
Minqing Hu and Bing Liu.
\newblock Mining and summarizing customer reviews.
\newblock In \emph{Proceedings of the Tenth ACM SIGKDD International Conference
  on Knowledge Discovery and Data Mining}, KDD '04, page 168–177, New York,
  NY, USA, 2004. Association for Computing Machinery.
\newblock ISBN 1581138881.
\newblock \doi{10.1145/1014052.1014073}.
\newblock URL \url{https://doi.org/10.1145/1014052.1014073}.

\bibitem[Socher et~al.(2013)Socher, Perelygin, Wu, Chuang, Manning, Ng, and
  Potts]{socher-etal-2013-recursive}
Richard Socher, Alex Perelygin, Jean Wu, Jason Chuang, Christopher~D. Manning,
  Andrew Ng, and Christopher Potts.
\newblock Recursive deep models for semantic compositionality over a sentiment
  treebank.
\newblock In \emph{Proceedings of the 2013 Conference on Empirical Methods in
  Natural Language Processing}, pages 1631--1642, Seattle, Washington, USA,
  October 2013. Association for Computational Linguistics.
\newblock URL \url{https://aclanthology.org/D13-1170}.

\bibitem[Lewis et~al.(2020)Lewis, Liu, Goyal, Ghazvininejad, Mohamed, Levy,
  Stoyanov, and Zettlemoyer]{lewis-etal-2020-bart}
Mike Lewis, Yinhan Liu, Naman Goyal, Marjan Ghazvininejad, Abdelrahman Mohamed,
  Omer Levy, Veselin Stoyanov, and Luke Zettlemoyer.
\newblock {BART}: Denoising sequence-to-sequence pre-training for natural
  language generation, translation, and comprehension.
\newblock In \emph{Proceedings of the 58th Annual Meeting of the Association
  for Computational Linguistics}, pages 7871--7880, Online, July 2020.
  Association for Computational Linguistics.
\newblock \doi{10.18653/v1/2020.acl-main.703}.
\newblock URL \url{https://aclanthology.org/2020.acl-main.703}.

\bibitem[Wang et~al.(2022{\natexlab{b}})Wang, Zhang, Wei, and
  Li]{wang2022focused}
Bin Wang, Chen Zhang, Chengwei Wei, and Haizhou Li.
\newblock A focused study on sequence length for dialogue summarization,
  2022{\natexlab{b}}.

\bibitem[Liu et~al.(2019)Liu, Ott, Goyal, Du, Joshi, Chen, Levy, Lewis,
  Zettlemoyer, and Stoyanov]{Liu2019RoBERTaAR}
Yinhan Liu, Myle Ott, Naman Goyal, Jingfei Du, Mandar Joshi, Danqi Chen, Omer
  Levy, Mike Lewis, Luke Zettlemoyer, and Veselin Stoyanov.
\newblock Roberta: A robustly optimized bert pretraining approach.
\newblock \emph{ArXiv}, abs/1907.11692, 2019.

\bibitem[Lundberg et~al.(2022)Lundberg, S{\'a}nchez~Vi{\~n}uela, and
  Biales]{lundberg-etal-2022-dialogue}
Conrad Lundberg, Leyre S{\'a}nchez~Vi{\~n}uela, and Siena Biales.
\newblock Dialogue summarization using {BART}.
\newblock In \emph{Proceedings of the 15th International Conference on Natural
  Language Generation: Generation Challenges}, pages 121--125, Waterville,
  Maine, USA and virtual meeting, July 2022. Association for Computational
  Linguistics.
\newblock URL \url{https://aclanthology.org/2022.inlg-genchal.17}.

\bibitem[Bhattacharjee et~al.(2022)Bhattacharjee, Shinde, Ghosal, and
  Ekbal]{bhattacharjee-etal-2022-multi}
Saprativa Bhattacharjee, Kartik Shinde, Tirthankar Ghosal, and Asif Ekbal.
\newblock A multi-task learning approach for summarization of dialogues.
\newblock In \emph{Proceedings of the 15th International Conference on Natural
  Language Generation: Generation Challenges}, pages 110--120, Waterville,
  Maine, USA and virtual meeting, July 2022. Association for Computational
  Linguistics.
\newblock URL \url{https://aclanthology.org/2022.inlg-genchal.16}.

\bibitem[Chauhan et~al.(2022)Chauhan, Roy, Dey, and
  Goel]{chauhan-etal-2022-tcs}
Vipul Chauhan, Prasenjeet Roy, Lipika Dey, and Tushar Goel.
\newblock {TCS}{\_}{WITM}{\_}2022 @ {D}ialog{S}um : Topic oriented
  summarization using transformer based encoder decoder model.
\newblock In \emph{Proceedings of the 15th International Conference on Natural
  Language Generation: Generation Challenges}, pages 104--109, Waterville,
  Maine, USA and virtual meeting, July 2022. Association for Computational
  Linguistics.
\newblock URL \url{https://aclanthology.org/2022.inlg-genchal.15}.

\end{thebibliography}

\section{Appendix A. Example Analysis}

\begin{figure*}[h!]
    \centering
    \includegraphics[width=\textwidth]{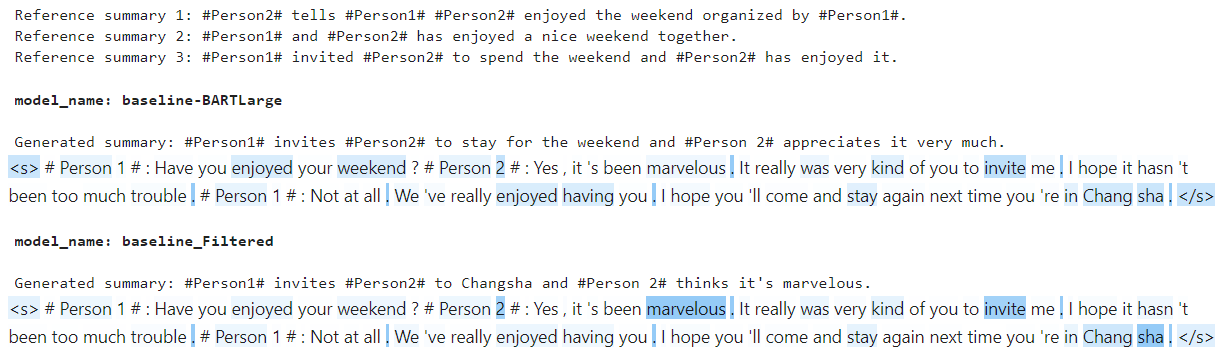}
    \caption{Example of test\_151, with three references, and predictions as well as visualization of the attention of two models: \baseline{} and \baselinefiltered{}.} %\_3 (number 455)
    \label{fig:test_151}
\end{figure*}

In addition to the previous example, we present two other examples in the following.
In Figure~\ref{fig:test_151}, attention is focused on words such as ``enjoyed", ``weekend", ``marvelous", ``kind", ``invite", ``enjoyed having", ``stay", ``Chang sha", which contain factual information or affective content. In the \baseline{} model, the words ``enjoyed" and ``enjoyed having" receive the most attention, and in \baselinefiltered{}, the word ``marvelous" attracts the most attention.

\begin{figure*}[h!]
    \centering
    \includegraphics[width=\textwidth]{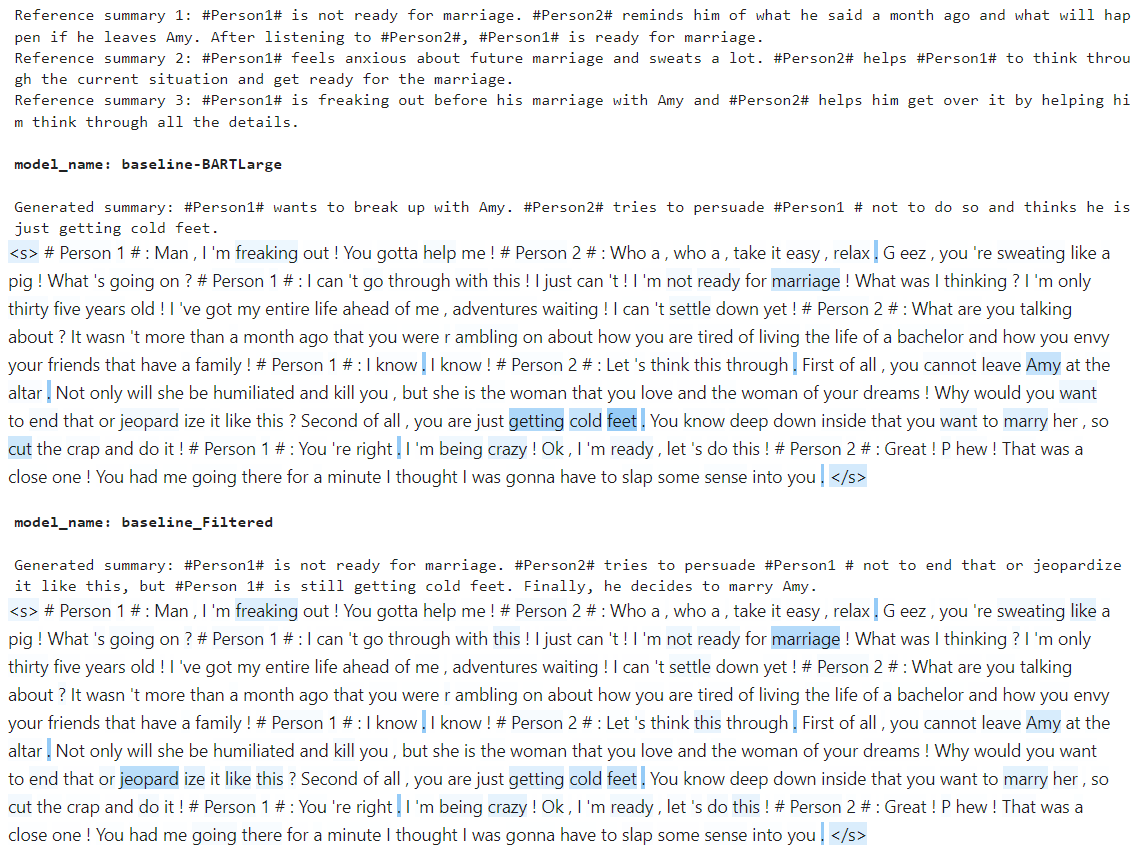}
    \caption{Example of test\_440, with three references, and predictions as well as visualization of the attention of two models: \baseline{} and \baselinefiltered{}.} %sample test\_440\_3 (number 1322)
    \label{fig:test_440}
\end{figure*}

In Figure~\ref{fig:test_440}, the expression “getting cold feet” is highlighted as a whole, indicating that the models have the ability to understand multi-word expressions. Furthermore, in \baselinefiltered{}, not only “freaking”, “marriage” and “being crazy” are highlighted, but also “jeopardize”.

\end{document}